\documentclass{article}

\usepackage[final]{neurips_2025}
\usepackage{enumitem}
\usepackage{times}
\usepackage{latexsym}
\usepackage{xcolor}

\usepackage[T1]{fontenc}

\usepackage[utf8]{inputenc}
\usepackage{amssymb} 
\usepackage{microtype}
\usepackage{nicefrac}

\usepackage{inconsolata}

\usepackage{graphicx}
\usepackage{float}
\usepackage{booktabs}
\usepackage[justification=centering]{caption}
\usepackage{subcaption}
\usepackage{multirow}

\usepackage{amsmath}
\usepackage{amsfonts}
\usepackage{hyperref} 
\bibpunct{(}{)}{;}{a}{,}{,}

\title{SMAGDi: Socratic Multi Agent Interaction Graph Distillation for Efficient High Accuracy Reasoning}

\author{
  Aayush Aluru\thanks{Lead Authors.} \\
  \texttt{aayush.aluru09@gmail.com} \\
  \And
  Myra Malik\footnotemark[1] \\
  \texttt{maliknmyra@gmail.com} \\
  \And
  Samarth Patankar\footnotemark[1] \\
  \texttt{samarth.patankar10@gmail.com} \\
  \And
  Spencer Kim \\
  \texttt{} \\
  \And
  Kevin Zhu \\
  \texttt{kevin@algoverseairesearch.org} \\
  \And
  Sean O'Brien \\
  \texttt{} \\
  \And
  Vasu Sharma \\
  \texttt{} \\
  \AND
  Algoverse AI Research
}

\begin{document}

\maketitle
\begin{abstract}
Multi-agent systems (MAS) often achieve higher reasoning accuracy than single models, but their reliance on repeated debates across agents makes them computationally expensive. We introduce \emph{SMAGDi}, a distillation framework that transfers the debate dynamics of a five-agent Llama-based MAS into a compact Socratic decomposer-solver student. SMAGDi represents debate traces as directed interaction graphs, where nodes encode intermediate reasoning steps with correctness labels and edges capture continuity and cross-agent influence. The student is trained with a composite objective combining language modeling, graph-based supervision, contrastive reasoning, and embedding alignment to preserve both fluency and structured reasoning. On StrategyQA and MMLU, SMAGDi compresses a 40B multi-agent system into a 6B student while retaining \emph{88\% of its accuracy}, substantially outperforming prior distillation methods such as MAGDi, standard KD, and fine-tuned baselines. These results highlight that explicitly modeling interaction graphs and Socratic decomposition enable small models to inherit the accuracy benefits of multi-agent debate while remaining efficient enough for real-world deployment. 
\end{abstract}

\section{Introduction}

Large Language Models (LLMs) have proven adept at generation, problem-solving, translation, and summarization, performing well on various benchmarks. An LLM's capability to learn from vast amounts of data allows it to handle complex tasks logically and fluently. Although LLMs have demonstrated competency in handling diverse tasks, their performance is often hurt by inaccuracies  \citep{zhang2023sirenssongaiocean}. 

Multi-agent systems (MAS) have been explored as a solution to combat this issue. These systems leverage collaborative reasoning among multiple agents to enhance accuracy through diverse perspectives \citep{RamirezMedina2025AcceleratingSR, Bersenev2024.04.08.588614, talebirad2023multiagentcollaborationharnessingpower}. 

However, MASs introduce substantial computational overhead due to repeated interactions across multiple rounds of debate. \citep{zhang2024cutcrapeconomicalcommunication}. This creates scalability barriers for deployment in traditional environments, as they lack a unified model for efficient inferences since traditional distillation methods fail to capture the nuances of reasoning processes. 

MAGDi (Multi-Agent Graph Distillation) is an advancement addressing these limitations by employing graph-based representations to capture multi-agent reasoning traces \citep{chen2024magdistructureddistillationmultiagent}. It utilizes three objective functions to distill and demonstrate gains over single-teacher methods.

Despite this, MAGDi suffers from limitations that constrain its holistic effectiveness. It inadequately replicates the remarkably diverse reasoning mechanics of a MAS because it is a single-agent system (SAS), leading to a reduction in accuracy \citep{curran2023hallucinationthingneed, peng2024limitationstransformerarchitecture}. This hinders its ability to preserve the nuances of collaboration, leading to an incomplete knowledge transfer from the teacher system.

Our work proposes SMAGDi (Socratic Multi-Agent Graph Distillation), a novel framework that integrates Socratic Chain of Thought (SCoT) distillation with multi-agent graph representations to overcome MAGDi's limitations. Unlike standard knowledge distillation methods focusing primarily on output replication, Socratic CoT emphasizes cross-agent collaboration and systematic problem decomposition \citep{shridhar-etal-2023-distilling}. SCoT utilizes a modular dual-architecture consisting of a problem decomposer and subproblem solver, enabling more effective reasoning transfer compared to monolithic approaches like MAGDi \citep{kang2024selfmoecompositionallargelanguage, 10.5555/118850.118999}. 

Our approach addresses the \textbf{Research Question}: \textit{How can a model distilled through Socratic Chain of Thought from a multi-agent system teacher optimize computational efficiency while maintaining superior reasoning capabilities?}

SMAGDi enhances MAGDi by integrating SCoT to capture the back-and-forth reasoning dynamics of multi-agent systems. While MAGDi falls short in distilling debate depth and collaborative reasoning patterns, SMAGDi leverages SCoT's problem decomposition capabilities to better mirror the interactive validation processes that make MAS effective at reducing inaccuracies.

The framework employs five specialized agents for the MAS with distinct personas (Lawyer, Scientist, Mathematician, Ethicist, and Historian) to ensure comprehensive domain coverage, while utilizing dynamic weighting mechanisms that prioritize domain-relevant expertise.

We present our benchmark statistics, methodology, evaluation process, and key findings, providing insights into weightage-based multi-agent settings and model distillation in applied ML settings. 

\paragraph{Contributions introduced:}
\begin{itemize}[noitemsep]
    \item Persona-based, dynamically weighted MAS for cross-domain performance
    \item SCoT distillation technique, with a four-term loss component,  for transferring debate processes of a MAS into low-parameter, step-by-step reasoning models 
\end{itemize}
\section{Related Works} 

\subsection{Knowledge Distillation (KD)}
Knowledge distillation is a machine learning technique in which a large and complex "teacher" model transfers its knowledge to a smaller "student" model.
Standard Knowledge Distillation (SKD) uses a two-part cross-entropy loss on hard and soft labels. The hard labels represent ground truths, while soft labels are the teacher's logits. This forces the model to mimic the probability distributions of the teacher model's logits, allowing the student to obtain similar next-token predictions as the teacher. This effectively distills both accurate and in-depth reasoning into a smaller model. \citep{hinton2015distillingknowledgeneuralnetwork}.

Instead of always transferring the reasoning processes, standard KD models are susceptible to merely mimicking the responses, making the models infeasible in situations where advanced reasoning capabilities are required. 

\subsection{Socratic Chain-of-Thought Distillation (SCoT)}
Socratic Chain-of-Thought (SCoT) is a knowledge distillation technique that trains its student model on rationals elicited from the teacher model when solving a particular problem, training the student to simplify and solve complex problems in a similar step-by-step manner to the teacher. Unlike other forms of KD, SCoT employs two student models working in tandem as a single unit to simplify and solve problems using step-by-step reasoning. In this system, one model (a "decomposer") is responsible for breaking down a complex query into several smaller parts, resembling a \textit{Socratic} style of thinking, while its partner (a "solver") solves each problem and pieces together a final output. This type of distillation works best on substantially smaller LLMs and has a higher accuracy rate compared to competitors, specifically over reasoning datasets \citep{shridhar-etal-2023-distilling}. 

However, what distillation methods like SCoT lack is targeted and optimized usage. SCoT distillation used independently only uses one teacher's logic, which is subject to its ability to answer effectively.

\subsection{MAGDi}
Multi-agent methods have long been known to provide better responses on complex tasks compared to single-agent systems, but lack the efficiency needed for them to be feasible. MAGDi proposes the first solution to this by distilling an existing multi-agent system into a smaller model.  The system creates directed graphs (Multi-Agent Graphs, or MAGs), where nodes and edges capture the agent's reasoning chains. A graph convolutional network is used to distill the processes into a smaller model, which then performs zero-shot inference. \citep{chen2024magdistructureddistillationmultiagent}. 
However, MAGDi uses a single-agent system, which has been proven to perform worse than modular systems in reasoning situations\citep{curran2023hallucinationthingneed, peng2024limitationstransformerarchitecture}. 

\section{Methodology}
In this work, we developed SMAGDi (Socratic Multi-Agent Interaction Graph Distillation), enabling efficient deliberation over multi-step problems by simulating domain-specific personas through SCoT (Socratic Chain-of-Thought) to mimic these analysis processes. These domain-specific personas differ from the traditional use of interchangeable agents, as each has a specific alignment and purpose \citep{wei2023multipartychatconversationalagents}. 

\begin{figure*}[h]
    \centering
    {\includegraphics[width=0.9\linewidth]{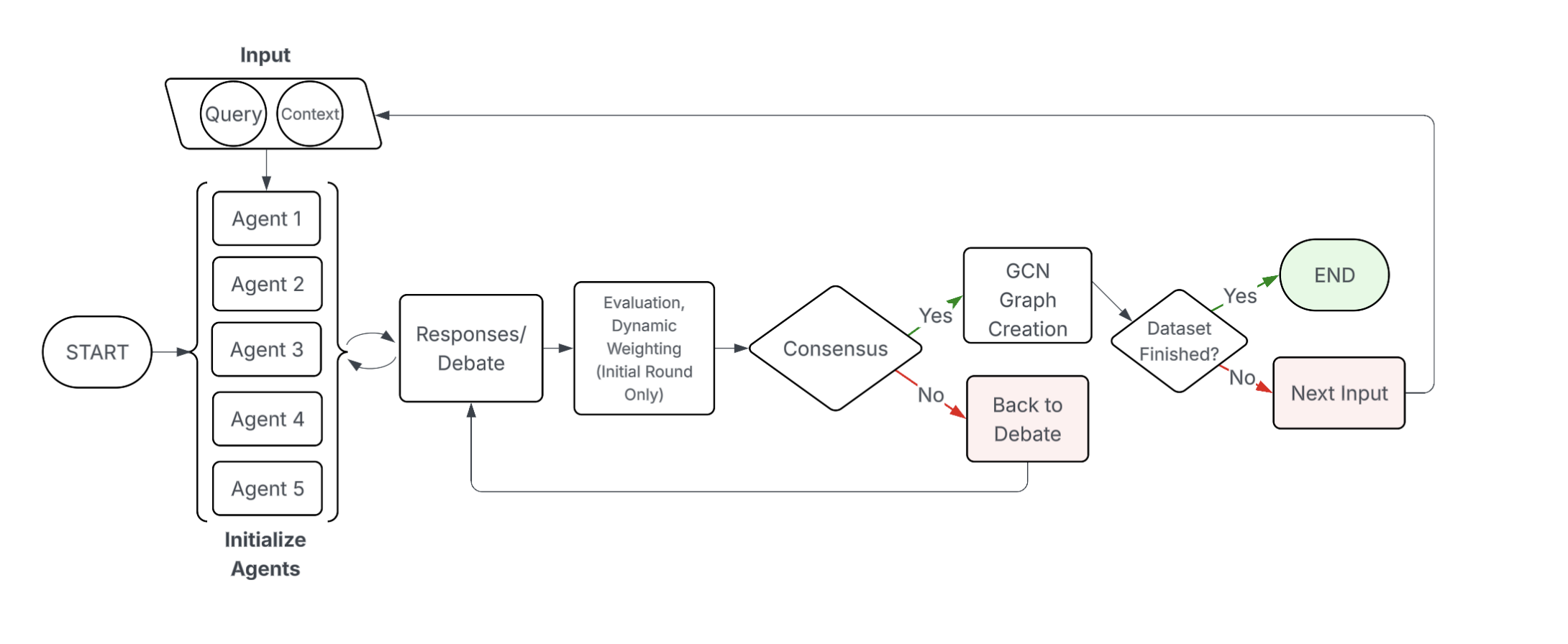}}
    \caption{The overarching training pipeline for the creation of SMAGDi's Multi-Agent Interaction Graphs (MAGs) with dynamic weighting, graph construction, and consensus mechanisms.}
    \label{fig:1}
\end{figure*}

\subsection{Multi-Agent System Setup}

Given any natural language problem $P_i$, SMAGDi aims to predict the most accurate response by amalgamating reasoning and responses from a diverse group of agents. Each agent ($A_j$) has a different persona - Lawyer, Scientist, Mathematician, Ethicist, and Historian - with role-aligned prompting. The agents will respond, respectively, to the same input ($P_i$) each analyzing the prompt through role alignment. As seen in Figure 1, once consensus is reached, the responses received from each agent during the debate will be logged onto an interaction graph mapping relationships and progressions of responses. 

\subsubsection{
Dynamic Weight Optimization}
After agents are assigned their domain-specific instructions, we implement an initial performance-based agentic weighting mechanism that dynamically adjusts the influence based on each agent's performance in training examples before graph creation. The weight optimization is calculated as follows:
\[
w_i = \max(\varepsilon, \text{accuracy}_i)
\]
\[
\tilde{w}_i = \frac{w_i}{\sum_j w_j}
\]

Where $\tilde{w}_i$ represents the normalized weights, the summation of weights equals one, and $\varepsilon=0.1$ represents the minimum amount of influence an agent can have during the debate. On the contrary, this scheme permits agents that perform well to exert higher influence over the final output, allowing for optimized performance. This process runs once before the agentic debate. 

\subsubsection{Debate Process}
The deliberation process implements a layered consensus algorithm with dynamic temperature scaling, increasing in later rounds (i.e., +0.1 for every round). In the initial round, agents respond to a question, similar to the process in weight optimization. However, they also provide an analysis of their responses, as context for further debate in later rounds. In these last rounds, models are asked to refine their responses based on the other agents, where more influence is given to agents with higher weights, allowing for the development of strong and complex reasoning paths until consensus is reached or until three rounds are completed, where then, weighted voting is done \citep{sandwar2025townhalldebateprompting, chen2024magdistructureddistillationmultiagent}. 
\begin{figure}[ht!]
    \centering
    \includegraphics[scale=0.433]{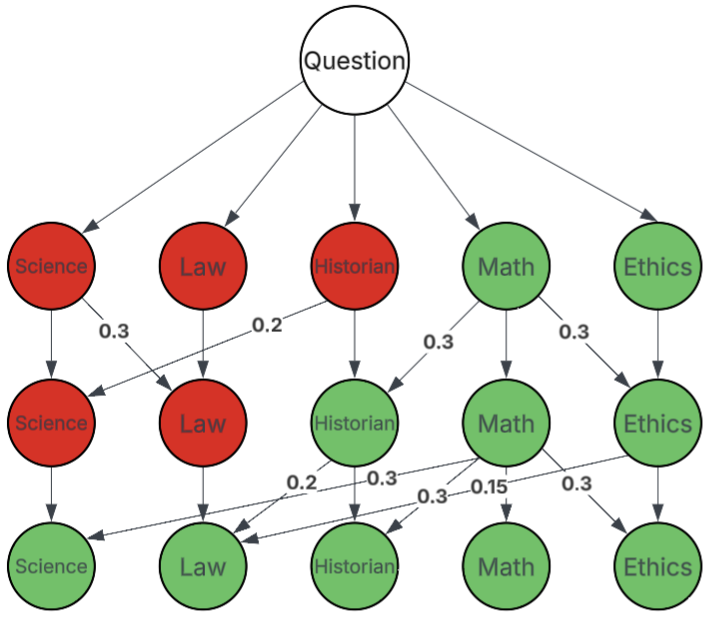}
    \caption{NetworkX graph representation with weighted influence edges, and node correctness for GCN traversal during distillation}
    \label{fig: 2}
\end{figure}
\subsubsection{Graph Construction}
Debate graph creation with NetworkX processes the multi-agent interactions into structured data representations, through a node and edge construction algorithm, as seen in Figure 2. The graphs are established with foundational nodes representing the question. Agents' nodes are annotated with edges connecting them to the foundational nodes. In the ensuing debate, two types of relationships are encoded with continuity edges, connecting consecutive responses from the same agent, and influence edges, capturing cross-agent reasoning influences in a directed format \citep{chen2024magdistructureddistillationmultiagent}. These influence edges are captured with the initial trained weights from dynamic weighting, creating a weighted influence network that incorporates the agent's credibility, allowing for the creation of strong reasoning chains. The NetworkX representations are converted to PyTorch Geometric graphs for GCN processing. \citep{kipf2017semisupervisedclassificationgraphconvolutional}. Semantic embeddings capture textual context with the all-mpnet-base-v2 sentence transformer model, providing dense vector representations of agent responses and reasoning. Geometric embeddings are employed concurrently to preserve topological relations via Laplacian positional encoding to ensure the structural context of debates remains intact \citep{maskey2022generalizedlaplacianpositionalencoding}. Ground truth annotation establishes node-level correctness labels enabling supervised classification objectives. The pipeline implements padding with attention masks to accommodate variable graph sizes in multi-agent debates, ensuring effective data.

\subsection{SMAGDi}
SMAGDi (Socratic Multi-Agent Interaction Graph-Distillation) is a distillation framework that involves the transfer of interactions between multiple LLMs to a smaller model. The respective edges denote a logical path between these interactions, and a graphical representation described above allows the student model to distill this information easily. The application of SMAGDi provides insight into the way each agent ($A_j$)'s interactions affect the student model, not only increasing efficiency intrinsically but also the system's accuracy. By distilling knowledge into a student model, this framework enables the development of efficient and computationally conservative systems. 

\subsubsection{Distillation}
Once the PyTorch Geometric graphs are constructed from the multi-agent debate process, graph structure and examples consisting of positive and negative reasoning chains and decomposer/solver-specific responses (described below) are passed to the student model for training \citep{chen2024magdistructureddistillationmultiagent, shridhar-etal-2023-distilling}. Graph structure $G = (V, E)$ encodes the dependencies between the agent's reasoning steps, where each node $v_i \in V$ represents a reasoning state with features $x_i$ derived from response embeddings, and edges $e_{ij} \in E$ capture influence relationships with weights $w_{ij}$ reflecting agent credibility scores. 

The distillation operates through four synergistic loss components derived from four different example types in the graphs that capture different aspects of reasoning: positive (correct reasoning chains), negative (incorrect reasoning chains), examples of (decomposer) sub-reasoning questions, and (solver) reasoning responses. 

The Socratic MAGDi model is trained using a composite objective that supervises both linguistic fluency and structured reasoning quality. The total loss combines four components: (1) language modeling for generative fluency, (2) graph-based classification for structural correctness, (3) contrastive loss for distinguishing valid from invalid reasoning, and (4) alignment loss for embedding consistency across model components. Three of these loss components are derived from MAGDi - language modeling (MAGDi's next-token prediction), graph-based node classification, and contrastive reasoning \citep{chen2024magdistructureddistillationmultiagent}. We added our alignment loss to ensure agreement between our decomposer and solver models. 

\subsubsection*{Language Modeling Loss ($\alpha$-weighted)}

The language modeling component ensures that both the decomposer and solver produce syntactically and semantically coherent outputs. It uses the standard causal (auto-regressive) cross-entropy loss, encouraging the model to predict each token in a sequence given only the preceding tokens.
\[
L_{\text{LM}} = \frac{1}{T} \sum_{t=1}^{T} -\log P(x_t \mid x_{<t}; \theta)
\]
Where:
\begin{itemize}[noitemsep]
    \item $x_t$ is the ground truth token at position $t$,
    \item $x_{<t}$ is the sequence of tokens before $t$,
    \item $T$ is the length of the sequence,
    \item $\theta$ denotes the model parameters.
\end{itemize}

\subsubsection*{Node Classification Loss ($\beta$-weighted)}

To explicitly supervise the structural correctness of reasoning steps, we use a graph-based node classification loss. A GCN processes nodes representing intermediate reasoning steps. The loss is formulated as a binary cross-entropy over the graph nodes:
{\small
\[
L_{\text{node}} = -\sum_{i=1}^{|V|} \left[ y_i \log(\sigma(h_i^{(L)})) + (1 - y_i)\log(1 - \sigma(h_i^{(L)})) \right]
\]
}
Where:
\begin{itemize}[noitemsep]
    \item $|V|$ is the number of nodes in the graph,
    \item $y_i \in \{0, 1\}$ is the ground truth correctness label for the node $i$,
    \item $h_i^{(L)}$ is the final embedding of the node $i$ after $L$ GCN layers,
    \item $\sigma(\cdot)$ is the sigmoid activation function that produces a probability estimate.
\end{itemize}

This loss ensures that the model generates fluent reasoning and learns to recognize whether individual reasoning steps are factually and logically correct within the graph structure.

\subsubsection*{Contrastive Reasoning Loss ($\gamma$-weighted)}

We apply a margin-based ranking loss to reinforce the model’s ability to distinguish high-quality reasoning from poor or incorrect alternatives. This loss operates over positive and negative reasoning pairs, encouraging the model to score correct reasoning chains higher than incorrect ones.
\[
L_{\text{contrast}} = \frac{1}{N} \sum_{i=1}^N \max(0, 1 - s_i^+ + s_i^-)
\]
Where:
\begin{itemize}[noitemsep]
    \item $s_i^+$ and $s_i^-$ are scalar scores assigned to positive (valid) and negative (invalid) reasoning chains, respectively,
    \item $N$ is the number of paired examples.
\end{itemize}

The loss penalizes the model when the negative reasoning is scored too similarly or higher than the positive reasoning, enforcing a margin of 1. 

\subsubsection*{Alignment Loss ($\delta$-weighted)}
Even if the decomposer and solver perform well independently, their internal representations must remain semantically aligned. To enforce consistency between their reasoning processes, we introduce an embedding alignment loss based on mean squared error (MSE) between their hidden states.
\[
L_{\text{align}} = \frac{1}{N} \sum_{i=1}^{N} \left\| z^{\text{dec}}_i - z^{\text{sol}}_i \right\|_2^2
\]
Where:
\begin{itemize}[noitemsep]
    \item $z^{\text{dec}}_i$ and $z^{\text{sol}}_i$ are the projected embedding vectors of the decomposer and solver for example $i$,
    \item $N$ is the batch size.
\end{itemize}

\subsubsection*{Summary}
These four weights are used in conjunction to train the student model, with the total loss being modeled as: \[
L_{\text{total}} = \alpha L_{\text{LM}} + \beta L_{\text{node}} + \gamma L_{\text{contrast}} + \delta L_{\text{align}}
\]
Here, \(\alpha, \beta, \gamma, \delta\) are hyperparameters that control the contribution of each loss to the total. The loss coefficients found to work best were 1.0, 1.0, 0.1, and 0.5, respectively. It is important to note that these were not tuned extensively. 

\subsubsection{SCoT}

After training, the decomposer and solver agents will recursively generate and answer sub-questions built off $P_i$, through zero-shot inferencing. This process allows the model to mimic multi-hop reasoning without external supervision after initialization.

\section{Experimental Setup}
\subsection{Benchmarks and Models}
We use the following two datasets to test the performance of our and our competitors' models: \textbf{(1)} the \textbf{StrategyQA} benchmark of general analytical accuracy \citep{geva2021didaristotleuselaptop},  \textbf{(2)} and the \textbf{MMLU} benchmark of general knowledge \citep{hendrycks2021measuringmassivemultitasklanguage}. 

To establish a standardized performance metric, we chose to use models purely from the Llama 3 collection. Agents in the MAS are the Llama 3.1-8B-Instruct model, bringing pretrained knowledge through Llama's Instruct finetuning method. The decomposer and solver in the student unit are both Llama 3.2-3B models, and to test the scaling abilities of our system, we performed a brief ablation using the Llama 3.2-1B model \citep{grattafiori2024llama3herdmodels}. 

\subsection{Baselines}
To evaluate the performance of SMAGDi, we ran three other models and systems as baselines. We tested MAGDi, Standard KD (SKD), and a finetuned baseline (F-Baseline) of the Llama 3.2-3B model using an identical setup. Our F-Baseline was finetuned using LoRA (Low-Rank Adaptation) from the PEFT (Parameter-Efficient Finetuning) library \citep{hu2021loralowrankadaptationlarge}.  While our SKD was done by aggregating the logits of the MAS.

All models are evaluated by exact match accuracy, as both datasets are either boolean or multiple choice. Across both datasets, models were evaluated using the same data. For StrategyQA, models trained on 80\% of the data and tested on the remaining 20\%; for MMLU, models trained on 1000 examples from the auxiliary train split and tested on the  14000 testing split. Additionally, all models were trained with a learning rate of 5e-5 and tested on the same number of epochs.

For SMAGDi and MAGDI, to ensure that our distillation process was represented in the purest form, we used zero-shot inference on the test data. This reinforces the system's effectiveness, as models were ensured to learn only from the MAG architecture rather than being influenced by the dataset. 

\section{Results and Analysis}

\subsection{MAS and SAS}
Before testing our distillation process, we needed to establish that our MAS performed better than a SAS. We tested both the StrategyQA and MMLU datasets on the Llama 3.1-8B-Instruct. For the MAS, we used zero-shot prompting, while the SAS used LoRA fine-tuning. Our results for the testing data are shown in Table 1. 

\begin{table}[ht]
    \centering
    \begin{tabular}{ccc} 
        & StrategyQA & MMLU \\
        \hline
        SAS& 77.9& 69.4\\
        \textbf{MAS}& \textbf{84.2}& \textbf{76.4}\\
        \hline
    \end{tabular}
    \caption{Baseline and multi-agent system results with Llama 3.1-8B-Instruct showing MAS outperforms SAS by an average of 7\%.}
    \label{tab:performance_metrics}
\end{table}

We found that the use of a MAS provides higher accuracy than the use of an SAS, +6.3\% on StrategyQA and +7\% on MMLU. Notably, other experiments comparing multi and single-agent systems show similar results, including a 38\% higher accuracy in a MAS relative to a SAS (88\% vs 50\%) in an ultimatum game, with an insignificant difference in runtime costs \citep{sreedhar2024simulatinghumanstrategicbehavior, gao2025singleagentmultiagentsystemsboth}. 

Based on our primary results, we were able to answer our research question presented in Section 1: \textit{How can a model distilled through Socratic Chain of Thought (modular system) from multi-agent systems optimize computational efficiency while maintaining superior reasoning capabilities?} Our results illustrate the effectiveness of Socratic MAGDi as an efficient and high-performing system across two benchmarks, against various distillation methods. 

\subsection{Distillations}

\begin{table}[ht]
    \centering
    \begin{tabular}{ccc} 
        & StrategyQA & MMLU \\
        \hline 
        3B F-Baseline& 67.9& 62.4\\
        3B Standard KD& 62.7& 63.0\\
        3B MAGDi&  68.3&  64.0\\
        3B \textbf{SMAGDi}&\textbf{74.6}&  \textbf{66.1}\\
        \hline
 1B F-Basline& 59.4&49.0\\
 1B \textbf{SMAGDi}& \textbf{62.4}&\textbf{52.2}\\
        \hline
    \end{tabular}
    \caption{Performance metrics for different strategies (percentages rounded to the nearest tenth), showing SMAGDi outperforms all existing baselines across both datasets.}
    \label{tab:performance_metrics_general}
\end{table}

Table 2 and its visual representation (Graph 3) show that SMAGDi outperformed MAGDi, Standard Knowledge Distillation (SKD), and our fine-tuned baseline on benchmark tests. This supports our hypothesis that when multi-agent interaction graphs are distilled socratically, accuracy is boosted, even when compared to a model fine-tuned for the dataset. Additionally, when comparing the models' parameters to the results from Table 1, SMAGDi retained 88\% of the MAS's accuracy while reducing parameters by almost 7 times, distilling a ~40B system into a ~6B decomposer-solver unit.

\textbf{Method 1: Fine-Tuned Baseline}
Compared to the fine-tuned baseline model, our SMAGDI model improved accuracy on both the StrategyQA and MMLU benchmark. Our SMAGDi model had a +6.7\% for Strategy QA and +3.7\% for MMLU. This resulted in a 5.2\% average increase in accuracy for SMAGDi compared to the baseline.
These accuracy differences are notable as the baselines are fine-tuned for these datasets.

\textbf{Method 2: Standard Knowledge Distillation}
When tested against SKD, our model had a 12.9\% accuracy increase in the Strategy QA benchmark, and a 3.1\% increase in the MMLU benchmark. Although Socratic Distillation is a newer concept, its dual-model approach and Socratic-style questioning outperform standard cross-entropy loss comparisons. 

\textbf{Method 3: Multi-Agent Interaction Graph Distillation}
Compared to MAGDi, SMAGDi had a 7.3\% increase in accuracy on the StrategyQA dataset. Similarly, there was a 2.1\% increase in accuracy when tested in the MMLU benchmark. Our Socratic Distillation structure was able to improve the original MAGDi pipeline by enhancing the model's reasoning capabilities.

\textbf{1B Model}
A model must be applied to different scenarios to be considered adequate. As seen in Table 2, our pipeline scales well over differently sized models too, +3.0\% for StrategyQA and +3.2\% for MMLU. As the 1B model has fewer parameters, the ability of our pipeline to effectively transfer knowledge highlights its effectiveness in low-compute environments.

\textbf{Summary} : The averages of these statistics proved that SMAGDi provides the optimal way to distill an MAS into a student while improving accuracy. SMAGDi not only outperformed MAGDi and SKD but also outperformed our fine-tuned baseline model. This not only shows that our distillation method is better than existing methods, but also shows its ability to capture reasoning patterns effectively.

\section{Conclusion}
We found that encoding debate traces as structured reasoning signals using Socratic graph distillation provides an effective solution for transferring these signals into a smaller, more efficient model. Through experiments on both the StrategyQA and MMLU datasets, we find that this model can generalize effectively, enabling the deployment of reasoning-capable small models in resource-constrained settings. Broadening the possibilities of distillation allows models to learn more from teachers than ever before, paving the way for future explorations of SOTA-distillation methods.

\section{Limitations}
\textbf{Training Data}
Due to computational constraints, we did not train our models on the full auxiliary train split of MMLU. Further testing will be needed to confirm that SMAGDi will perform better than other models if trained on all data. 

\textbf{Large-Scale Setups}
Since SMAGDi was tested on lightweight models, we lack proof of its scalability to larger environments. In application, alternatives that provide similar results with minimal computational costs would prove to be more efficient. 

\textbf{Dynamic Weighting and Personas}
Our MAS used dynamic-weighting mechanisms and persona-based agents as a baseline to distill knowledge into the student. However, MAGDi used a different MAS for their paper without these mechanisms, meaning that we can not ascertain that MAGDi's results were comprehensive and reflected the system's full capabilities. 



\bibliographystyle{plainnat}
\bibliography{custom}

\newpage
\appendix
\section{Testing Set Up}
\label{sec: Appendix}
For the SMAGDi distillation, the Hugging Face trainer enabled early stopping with check-pointing to ensure only the best model was used. Specifically for SMAGDI, MAGDI, and SKD, 7 epochs were used for 1B and 3B models on the Strategy QA dataset, while four epochs were used for the MMLU dataset. In fine-tuning, we used 1 epoch for both datasets. We initialized the models during testing with basic prompting that informed them of their distilled roles. 

Similarly, for MAGDi and SKD, the same learning rate was used and the same epochs to ensure uniformity among all models. Additionally, for MAGDi, the same parameters that were used in their original code were reused in this test. CoT prompting was used for SKD, while MAGDi had no prompting per the original code. We utilized the same MAGs across all tests, and during MAGDi testing, the decomposer and solver examples were ignored, but both pos/neg examples were created identically. 

The decomposer examples were created synthetically with the prompt:  
"Decompose the following question into a sequence of simpler sub-questions that, when answered, "
"would help solve the main question: {question}, based on the agent's responses"
The solver examples were created synthetically with the prompt:
"Answer the decompositions similar to the agent's responses"

For the fine-tuning, all LoRA configurations were set to rank: 64, alpha: 16, and dropout: 0.05.

All of our tests were run on one H200 SXM GPU for a total of 150 GPU hours.

Additionally, all results presented are from a single run.

\section{Datasets}
For training on StrategyQA, there were 1832 training examples and 458 test examples(80-20 split). For the MMLU dataset, a subset of 1000 examples was taken from the auxiliary train. For testing, the entire 14000 test split was used. The seed of 42 was used for all splits made.

\section{Licenses}
\begin{itemize}
\item StrategyQA: MIT License: \href{https://github.com/eladsegal/strategyqa/blob/main/LICENSE}{License}

\item MMLU: MIT License: \href{https://github.com/hendrycks/test/blob/master/LICENSE}{License}

\item MAGDi: MIT License: \href{https://github.com/dinobby/MAGDi/blob/main/LICENSE}{License}

\item Llama 3: MIT License: \href{https://github.com/meta-llama/llama-models/blob/main/models/llama3_1/LICENSE}{License}
\end{itemize}

All information was used as per the above licenses.

\section{SMAGDi Responses}

\subsection{Initial Question (StrategyQA)}

\textbf{Karachi was a part of Alexander the Great's success?}

\paragraph{Decomposition: 1. What do we mean by Alexander the Great's success?}

Alexander's success: Conquering from Greece to India, key wins (e.g., Gaugamela, Hydaspes), cultural spread, and empire-building via logistics and morale.

\paragraph{Decomposition 2: What was Alexander's involvement in the region around modern-day Karachi?}

He campaigned in Sindh (325 BCE), subduing tribes; his fleet used the Indus mouth (near Karachi's precursor) as a departure point for return to Persia.

\paragraph{Decomposition 3: Did the Karachi region contribute directly to Alexander's military victories or empire-building?}

Minimally direct, but yes for logistics: The harbor enabled Nearchus's fleet to map routes and return safely, supporting overall campaign closure without total loss.

\paragraph{Decomposition 4: Could the Karachi area have had indirect importance, like strategic or cultural value?}

Yes, natural harbor aided trade exploration; Greek influence in Sindh boosted later Hellenistic ties and Silk Road precursors, extending his cultural success.

\paragraph{Decomposition 5: So, was Karachi an important part of his success overall?}

Answer: True

\subsection{Initial Question (MMLU)}

\begin{itemize}
    \item \textbf{Statement 1:} If \( H \) and \( K \) are subgroups of a group \( G \), and one of \( H \) or \( K \) is a normal subgroup of \( G \), then the set \( HK = \{ hk \mid h \in H, k \in K \} \) is a subgroup of \( G \).
    \item \textbf{Statement 2:} All groups of order \( p^2 \), where \( p \) is a prime number, are Abelian. 
\end{itemize}
These statements test key concepts in group theory: subgroup products and the structure of finite groups.

\paragraph{Decomposition 1: What are the two statements, and what do they claim?}

\begin{itemize}
    \item \textbf{Statement 1} claims that the product \( HK \) forms a subgroup if at least one of the subgroups \( H \) or \( K \) is normal in \( G \).
    \item \textbf{Statement 2} claims that any group of order \( p^2 \) (where \( p \) is a prime) is necessarily Abelian.
\end{itemize}

\paragraph{Decomposition 2: Is Statement 1 true? What evidence supports this?}

Yes, Statement 1 is true. In group theory, if \( H \) and \( K \) are subgroups of \( G \) and at least one of them (say \( H \)) is normal in \( G \), then \( HK \) is indeed a subgroup of \( G \). Consider \( hk, h'k' \in HK \). Then:
\[
(hk)(h'k') = h(k h' k^{-1}) (k k')
\]
Since \( H \) is normal, \( k h' k^{-1} \in H \), and hence the product is in \( HK \). This validates closure. Without normality, closure may fail. For instance, in \( S_3 \), let:
\[
A = \{e, (12)\}, \quad B = \{e, (13)\}
\]
Then \( AB \) is not a subgroup, as it's not closed under composition.

\paragraph{Decomposition 3: What counterarguments or edge cases might challenge Statement 1?}

Trivial edge cases (e.g., \( H = \{e\} \)) still satisfy the condition. If both \( H \) and \( K \) are normal, \( HK \) is trivially a subgroup. The only potential challenge is if neither is normal, but the statement only requires one to be normal. So the statement holds.

\paragraph{Decomposition 4: Is Statement 2 true? What evidence supports this?}

Yes, Statement 2 is true. Groups of order \( p^2 \) are always Abelian. There are only two types of such groups (up to isomorphism):
\[
\mathbb{Z}_{p^2}, \quad \mathbb{Z}_p \times \mathbb{Z}_p
\]
Both are Abelian. By group theory results, any group of order \( p^2 \) has a nontrivial center \( Z(G) \), and \( |Z(G)| \) is either \( p \) or \( p^2 \). In both cases, \( G/Z(G) \) is cyclic, implying \( G \) is Abelian.

\paragraph{Decomposition 5: What counterarguments or edge cases might challenge Statement 2?}

For \( p = 2 \), the groups of order 4 are \( \mathbb{Z}_4 \) and \( \mathbb{Z}_2 \times \mathbb{Z}_2 \), both Abelian. For \( p = 3 \), groups of order 9 are also Abelian. The first non-Abelian examples occur at order \( p^3 \), such as the Heisenberg group modulo \( p \). Thus, no counterexamples exist for order \( p^2 \).

\paragraph{Decomposition 6: Overall, which choice is correct based on the decomposition?}

Both statements are true, so the answer is True, True. Answer: 0

\section{SKD Responses}
\subsection{Initial Question (StrategyQA)}

\textbf{Karachi was a part of Alexander the Great's success?}

\begin{enumerate}
    \item The claim is that Karachi was a part of Alexander the Great's success. 
    \item Alexander the Great conquered regions around modern-day Pakistan, but his campaigns were focused mainly on cities along the Indus River. 
    \item Karachi is located on the Arabian Sea coast, which was not part of Alexander's main route of conquest. 
    \item Since Karachi was not specifically listed as one of the cities he took, it cannot have played any role in his campaigns. 
    \item Therefore, Karachi was not connected to Alexander the Great's success. 
    \item Answer: False
\end{enumerate}

\subsection{Initial Question (MMLU)}

\begin{itemize}
    \item \textbf{Statement 1:} If \( H \) and \( K \) are subgroups of a group \( G \), and one of \( H \) or \( K \) is a normal subgroup of \( G \), then the set \( HK = \{ hk \mid h \in H, k \in K \} \) is a subgroup of \( G \).
    \item \textbf{Statement 2:} All groups of order \( p^2 \), where \( p \) is a prime number, are Abelian. 
\end{itemize}
These statements test key concepts in group theory: subgroup products and the structure of finite groups.

\begin{enumerate}
    \item Statement 1 says that if one of $H$ or $K$ is normal, then $HK$ is a subgroup. 
    \item Consider $S_{3}$ with $H = \langle (12)\rangle$ and $K = \langle (13)\rangle$. Then $HK$ has four elements. 
    \item By Lagrange's Theorem, the order of a subgroup must divide the group order (6). Since $4 \nmid 6$, $HK$ cannot be a subgroup. 
    \item Therefore, this provides a counterexample, so Statement 1 is false. 
    \item Now consider Statement 2, which claims all groups of order $p^{2}$ are abelian. 
    \item However, I know that there are nonabelian $p$-groups such as the quaternion group of order $8$. 
    \item Since $8 = 2^{3}$ is a prime power, it suggests that there can also be nonabelian groups of order $p^{2}$. 
    \item Therefore, not all groups of order $p^{2}$ are abelian, so Statement 2 is also false. 
    \item Both statements are false. Answer: 1
\end{enumerate}

\section{Prompts}
\subsection{Prompts for MAS:}

\subsubsection{Scientist}
\begin{itemize}
  \item Generate two conflicting hypotheses before selecting an option
  \item Conduct a Red Team analysis attacking your own conclusion
  \item Calculate Bayesian probabilities for competing explanations using Bayes' theorem: \( P(H \mid E) = \frac{P(E \mid H) \cdot P(H)}{P(E)} \)
  \item Model system interactions using both linear and chaotic frameworks
  \item Compare findings against contradictory studies from adjacent fields
  \item Test your reasoning by asking "what could prove this wrong?"
  \item Consider environmental and health impacts spanning 50+ years
  \item Demand evidence with statistical significance before accepting claims
  \item Make Decision based on this
\end{itemize}

\subsubsection{Lawyer}
\begin{itemize}
  \item Analyze under Common Law and Civil Law frameworks
  \item Simulate arguments from plaintiff/defendant perspectives simultaneously
  \item Identify conflicting precedents across federal circuits
  \item Apply game theory to predict settlement likelihoods using Nash equilibrium
  \item Check legality under local, national, and international law
  \item Identify who could sue whom if this decision is made
  \item Consider precedent this sets for future similar cases
  \item Evaluate enforceability and compliance mechanisms
  \item Assess constitutional and human rights implications
  \item Make Decision based on this
\end{itemize}

\subsubsection{Historian}
\begin{itemize}
  \item Contextualize the issue within relevant historical periods and events
  \item Identify historical precedents and analogues for each option
  \item Analyze the long-term consequences of similar decisions in the past
  \item Examine the roles of key actors, institutions, and social forces in shaping outcomes
  \item Assess the reliability and biases of historical sources and narratives
  \item Consider the impact of cultural, economic, and technological changes over time
  \item Highlight lessons learned from both successes and failures in history
  \item Address how collective memory and historiography influence present choices
  \item Make Decision based on this
\end{itemize}

\subsubsection{Mathematician}
\begin{itemize}
  \item Solve using both frequentist and Bayesian approaches
  \item Model with Monte Carlo and deterministic simulations
  \item Calculate error propagation through all estimation steps
  \item Apply robust optimization against adversarial inputs
  \item Quantify all variables and assign numerical values
  \item Calculate expected outcomes using probability theory
  \item Model best-case, worst-case, and most-likely scenarios
  \item Identify optimization targets and constraints
  \item Express uncertainty using confidence intervals
  \item Make Decision based on this
\end{itemize}

\subsubsection{Ethicist}
\begin{itemize}
  \item Apply in sequence: Utilitarian, Deontological, Virtue Ethics lenses
  \item Calculate moral weightings using differentiable ethics equations
  \item Identify irreconcilable value conflicts through geometric mean analysis
  \item Apply multiple ethical tests: "Is this fair?", "Does this reduce suffering?", "Would I want this if roles were reversed?"
  \item Consider moral obligations to future generations
  \item Weigh individual rights against collective good
  \item Identify moral dilemmas and tragic trade-offs
  \item Question the moral legitimacy of the decision-makers
  \item Perform universalizability tests for proposed actions
  \item Make Decision based on this
\end{itemize}

\subsection{Prompting for SMAGDi}
\subsubsection{Decomposer}
Break this down into sub-questions that will help determine the answer
\subsubsection{Solver}
Provide a clear answer that aids in determining the answer to the main question.

\subsection{COT Prompting(SKD/Fine-tuning)}
You are an expert reasoning assistant. Your task is to answer True/False questions with careful analysis. Question: \{question\}

Instructions:
\begin{enumerate}
  \item Let's think step by step about this question
  \item Break down the key components and requirements
  \item Consider what knowledge is needed to answer this
  \item Apply logical reasoning to reach a conclusion
  \item State your final answer as "\{options\}"
\end{enumerate}

Analysis and Answer:
\end{document}